\documentclass[letterpaper, 10 pt, conference]{ieeeconf}
\IEEEoverridecommandlockouts
\usepackage{cite}
\usepackage{amsmath,amssymb,amsfonts}
\usepackage{graphicx}
\usepackage{textcomp}
\usepackage{adjustbox}
\usepackage{balance}
\usepackage{threeparttable}
\usepackage{soul}
\usepackage{verbatim}
\usepackage{url}
\usepackage{amssymb}
\usepackage{mathtools}
\usepackage{xparse}
\usepackage[percent]{overpic}
\usepackage{textcomp}
\usepackage{comment}
\usepackage{multirow}
\usepackage{array}
\usepackage{nicefrac} 
\usepackage{algorithm} 
\usepackage{algpseudocode} 
\usepackage[hidelinks]{hyperref}
\hypersetup{
    colorlinks=false,
    linkcolor=blue,
    filecolor=none,      
    urlcolor=blue,
    pdftitle={Overleaf Example},
    pdfpagemode=FullScreen,
    }
\makeatletter
\newcommand{\raisemath}[1]{\mathpalette{\raiseMath{#1}}}%
\newcommand{\raiseMath}[3]{\raisebox{#1}[0pt][0pt]{$#2#3$}}
\makeatother

\NewDocumentCommand{\qbar}{O{0.5pt} O{-6.55pt}}{
	\ensuremath{\mathrlap{\raisemath{#2}{\hspace*{#1}{\mathchar'26\mkern-9mu}}} q}%
}

\NewDocumentCommand{\qbars}{O{0.5pt} O{-4.65pt}}{
	\ensuremath{\mathrlap{\raisemath{#2}{\hspace*{#1}{\mathchar'26\mkern-9mu}}} q}%
}

\NewDocumentCommand{\qbarc}{O{0.5pt} O{-5.2pt}}{
	\ensuremath{\mathrlap{\raisemath{#2}{\hspace*{#1}{\mathchar'26\mkern-9mu}}} q}%
}

\NewDocumentCommand{\pbar}{O{-1.5pt} O{-6.65pt}}{
	\ensuremath{\mathrlap{\raisemath{#2}{\hspace*{#1}{\mathchar'26\mkern-9mu}}} p}%
}

\newcommand{\bs}[1]{\boldsymbol{#1}}  
\newcommand{\ts}[1]{\text{#1}}

\renewcommand{\baselinestretch}{0.9265}  

\def\BibTeX{{\rm B\kern-.05em{\sc i\kern-.025em b}\kern-.08em
    T\kern-.1667em\lower.7ex\hbox{E}\kern-.125emX}}
\begin{document}

\title{\LARGE \bf MPS: A New Method for Selecting the Stable \mbox{Closed-Loop} Equilibrium \mbox{Attitude-Error} Quaternion of a UAV During Flight \\
\thanks{This work was supported in part by the Joint Center for Aerospace Technology Innovation (JCATI) through Award\,172, in part by the Washington State University (WSU) Foundation and the Palouse Club through a Cougar Cage Award to N. O. P\'erez-Arancibia, and in part by the WSU Voiland College of Engineering and Architecture through a start-up package to N.~O.~P\'erez-Arancibia.}
\thanks{F.~M.~F.~R.~Gon\c{c}alves, K. I. Matveev, and N.~O.~P\'erez-Arancibia are with the School of Mechanical and Materials Engineering, Washington State University (WSU), Pullman, WA 99164-2920, USA (\mbox{e-mail:} {\tt francisco.goncalves@wsu.edu} (F. M. F. R. Gon\c{c}alves); {\tt matveev@wsu.edu} (K. I. Matveev); {\tt n.perezarancibia@wsu.edu} (N. O. P\'erez-Arancibia)). R.~M.~Bena is with the Department of Aerospace and Mechanical Engineering, University of Southern California (USC), Los Angeles, CA 90089-1453, USA (e-mail: {\tt bena@usc.edu}).}%
}

\author{\mbox{Francisco M. F. R. Gon\c{c}alves}, Ryan M. Bena, Konstantin I. Matveev, and \mbox{N\'estor O. P\'erez-Arancibia}}

\maketitle
\thispagestyle{empty}
\pagestyle{empty}

\begin{abstract}
We present \textit{model predictive selection} (MPS), a new method for selecting the stable \textit{\mbox{closed-loop}} (CL) equilibrium \textit{\mbox{attitude-error} quaternion} (AEQ) of an \textit{uncrewed aerial vehicle} (UAV) during the execution of \mbox{high-speed} yaw maneuvers. In this approach, we minimize the cost of yawing measured with a \textit{performance figure of merit} (PFM) that takes into account both the \mbox{aerodynamic-torque} control input and \mbox{attitude-error} state of the UAV. Specifically, this method uses a control law with a term whose sign is dynamically switched in real time to select, between two options, the torque associated with the lesser cost of rotation as predicted by a dynamical model of the UAV derived from first principles. This problem is relevant because the selection of the stable CL equilibrium AEQ significantly impacts the performance of a UAV during \mbox{high-speed} rotational flight, from both the power and \mbox{control-error} perspectives. To test and demonstrate the functionality and performance of the proposed method, we present data collected during one hundred \mbox{real-time} \mbox{high-speed} \mbox{yaw-tracking} flight experiments. These results highlight the superior capabilities of the proposed \mbox{MPS-based} scheme when compared to a benchmark controller commonly used in aerial robotics, as the PFM used to quantify the cost of flight is reduced by $\bs{60.30\,\%}$, on average. To our best knowledge, these are the first \mbox{flight-test} results that thoroughly demonstrate, evaluate, and compare the performance of a \mbox{real-time} controller capable of selecting the stable CL equilibrium AEQ during operation.
\end{abstract}

\section{Introduction}
\label{SECTION1}
A significant number of schemes developed for controlling the attitude of spacecraft, \textit{vertical \mbox{take-off} and landing} (VTOL) \textit{uncrewed aerial vehicles} (UAVs), and microrobotic flyers use \mbox{quaternion-based} control \mbox{laws \cite{ Ying_ICRA_2016, Ying_ACC_2017, Ying_IROS_2018, Ying_ICRA_2019, Ying_IEEE_TCST_2020, Ying_Automatica_2024, calderon2019control, BenaMPPC2022, Bena2023Yaw,yang2019bee,Bena_RAL_Perception, Salcudean, Wie_Sign, Thienel_sign, Kristiansen, Fjellstad_sign, Tayebi, Carino_QQC, YANG2012198, Fresk_Fullquat, wie1989quaternion}}. Controllers based on the use of \textit{\mbox{attitude-error} quaternions} (AEQs) exhibit numerous advantages compared to control methods that use Euler angles and rotation matrices~\mbox{\cite{Perez2013ICAR,Perez2015ModelFree}}. In particular, quaternions eliminate singularity issues inherent to attitude representations that use Euler angles and provide a compact \mbox{four-element} storage format instead of the \mbox{nine-element} format of rotation matrices. These characteristics improve numerical robustness and decrease computational complexity, enabling faster sampling rates for \mbox{real-time} control~\cite{comparisonEulerQuaternion, wen1991attitude, EulerQuaternionEfficiency}. However, overlapping of the rotational \mbox{state-space} induces a \mbox{well-known} ambiguity problem characterized by the existence of two different symmetric quaternions that can be used to represent the same attitude of an object in space. In particular, we show in~\cite{BenaMPPC2022,Bena2023Yaw} that when \mbox{quaternion-based} attitude representations are used in a \textit{\mbox{closed-loop}} (CL) scheme developed to control \mbox{rigid-body} flyers, the resulting dynamical systems have two different equilibria; one asymptotically stable and another unstable, depending on the definiteness---positive or negative---of a \mbox{controller-gain} matrix. Therefore, the stability of these two equilibria can be interchanged in real time by simply multiplying a term in the control law by $-1$. It is important to note that, even though both possible stable CL equilibrium AEQs represent exactly the same control error, the rotational trajectories required to eliminate this error are different; one has exactly the same direction of the instantaneous Euler axis of the first AEQ whereas the other has the opposite direction of the instantaneous Euler axis of the first AEQ.

In the case discussed in~\cite{BenaMPPC2022,Bena2023Yaw}, the existence of two fixed points with opposite stability properties and the utilization of \mbox{real-time} switching do not represent a problem because if a trajectory were to start at the unstable equilibrium, any perturbation---large or small---would cause the system state to asymptotically converge to the stable one. A main reason for using stability switching is the simultaneous improvement of flight performance and power utilization. Specifically, this technique is commonly used to avoid unwinding~\cite{bhat2000topological}, a phenomenon characterized by an abrupt \mbox{$2\pi$-rad} rotation, about the AEQ's Euler axis, of the UAV's \mbox{body-fixed} frame during flight. Avoiding unwinding is important during the execution of tasks that require \mbox{high-performance} flight and synchronization, such as \mbox{high-speed} object tracking, precise positioning, and landing on rotating platforms. Power optimization is crucial in \mbox{long-lasting} operations, such as search and rescue, traffic control, and surveillance. 

The simplest automatic \mbox{stability-switching} technique can be implemented by including in the control law, which generates the torque inputted to the system, the sign of the scalar part of the AEQ~\cite{Salcudean, Wie_Sign, Thienel_sign, Kristiansen, Fjellstad_sign, Bena_RAL_Perception, Bena2023Yaw, BenaMPPC2022, yang2019bee}. In particular, by multiplying this signum function and a term proportional to the vector part of the AEQ in the definition of the control law \mbox{in~\cite{Bena2023Yaw, BenaMPPC2022}}, we ensure that this new resulting term generates a torque in the direction of the shorter rotational trajectory required to eliminate the attitude error. However, as discussed in~\cite{quatOptimal}, applying a torque in the direction of the shorter rotational trajectory required to eliminate the attitude error does not necessarily ensure the best flight performance because, depending on the current angular velocity and orientation of the UAV, it might be more advantageous to apply a torque in the direction of the longer rotational path. For example, \cite{quatOptimal} proposes a set of \textit{a-priori} rules for
selecting the direction of the instantaneous control torque
inputted during flight that decreases the cost of flight according to a heuristic quantification method. These rules were derived
from data obtained through the simulation of maneuvers with
different orientation and \mbox{angular-velocity} initial conditions. 

Also, the problem can be approached with the methods presented in~\cite{Mayhew_Robust}~and~\cite{QuatAutomatica}. The research presented in~\cite{Mayhew_Robust} developed two methods for synthesizing hybrid controllers aiming to minimize the cost of flight, respectively based on \mbox{energy-like} and backstepping Lyapunov functions, while \cite{QuatAutomatica} introduces a switching controller aiming to minimize the cost of flight, explicitly quantified by a function of the \mbox{angular-velocity} and \mbox{attitude-tracking} errors during operation. With the exception of the method in~\cite{quatOptimal}, which selects the CL equilibrium AEQ \mbox{off line} using trajectory information known \textit{a priori}, all other \mbox{CL-fixed-point} selection algorithms use instantaneous state information only. Here, we introduce an alternative approach based on \textit{model predictive selection} (MPS), according to which a \mbox{real-time} controller employs a dynamic model of the controlled UAV to estimate the system's state evolution and predict the most \mbox{cost-efficient} stable CL equilibrium AEQ over a \mbox{finite-time} horizon, according to a \textit{performance figure of merit} (PFM). The main contributions of this paper are: (i)~a new \mbox{model-predictive} method for the selection of the most \mbox{cost-efficient}, according to a chosen PFM, stable CL equilibrium AEQ over a \mbox{finite-time} horizon; and, to our best knowledge, (ii)~the presentation of the first \mbox{real-time} \mbox{flight-test} results that demonstrate the operation and performance of a controller that dynamically selects the stable CL equilibrium AEQ of a UAV during flight.

The rest of the paper is organized as follows. Section\,\ref{Section02} describes the kinematics and dynamics of the UAV used in the presented research, discusses the structure of a continuous controller used as the starting point in the analyses that follow, and presents the stability analysis of the two CL equilibrium AEQs of the system. Section\,\ref{Section03} describes the performance problem associated with the selection of the stable CL equilibrium AEQ and the controller used as benchmark to analyze the experimental results. Section\,\ref{Section04} introduces the proposed \mbox{MPS-based} controller. Section\,\ref{Section05} presents and discusses experimental results obtained using both the \mbox{MPS-based} and benchmark controllers. Last, Section\,\ref{Section06} states some conclusions regarding the presented research. 

\vspace{1ex}
\textit{\textbf{Notation:}}
\begin{enumerate}
\item Italic lowercase symbols represent scalars, e.g., $p$; bold lowercase symbols represent vectors, e.g., $\bs{p}$; bold uppercase symbols represent matrices, e.g., $\bs{P}$; and bold crossed lowercase symbols represent quaternions, \mbox{e.g., $\bs{\pbar}$}.
\item The real variable $t$ denotes continuous time. The integer variables $i$ and $k$ are used to index discrete time.
\item Differentiation with respect to time is denoted by the dot operator, e.g., $\dot{p} = \frac{dp}{dt}$. 
\item The symbol $\times$ denotes the \mbox{cross-product} of two vectors. Multiplication between two quaternions is denoted by $\otimes$.
\item The $2$-norm of a vector is denoted by the operator $\| \cdot \|_2$.
\item The transpose of a matrix is denoted by $\left[ \, \cdot \, \right]^T$.
\item The operator $\ts{sgn} \left\{ \cdot \right\}$ extracts the sign of a scalar.
\end{enumerate}
\begin{figure}[t!]
\vspace{1.2ex}
\begin{center}
\includegraphics[width=0.48\textwidth]{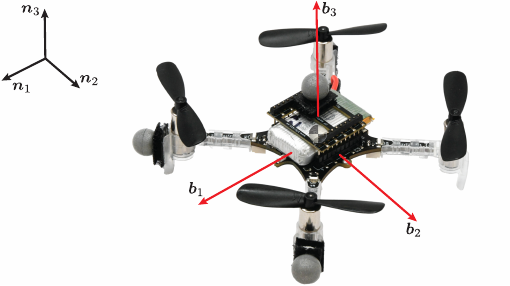}
\end{center}
\vspace{-2ex}     
\caption{\textbf{Photograph of the UAV platform used in the flight tests, the Crazyflie\,2.1, and depiction of the frames of reference used to describe its kinematics.} \mbox{$\bs{\mathcal{N}} = \{\bs{n}_1, \bs{n}_2, \bs{n}_3\}$} denotes the inertial frame of reference; \mbox{$\bs{\mathcal{B}} = \{\bs{b}_1, \bs{b}_2, \bs{b}_3\}$} denotes the \mbox{body-fixed} frame of reference, whose origin coincides with the center of mass of the UAV.
\label{Fig01}}
\vspace{-2ex}
\end{figure}

\section{Dynamic Model and Continuous Controller}
\label{Section02}
\subsection{Reduced-Complexity Rigid-Body Attitude Dynamics}\label{Subsection02A}
The UAV used in the research presented in this paper is shown in Fig.\,\ref{Fig01}. Also, this figure graphically defines the inertial and \mbox{body-fixed} frames, \mbox{$\bs{\mathcal{N}} = \{\bs{n}_1, \bs{n}_2, \bs{n}_3\}$} and \mbox{$\bs{\mathcal{B}} = \{\bs{b}_1, \bs{b}_2, \bs{b}_3\}$}, employed to describe the kinematics of the UAV during flight. The origin of $\bs{\mathcal{B}}$ coincides with the location of the robot's center of mass. Here, we focus only on the attitude dynamics of the flyer, which we represent using the model thoroughly discussed in \cite{Bena2023Yaw, BenaMPPC2022,calderon2019control, Ying_ACC_2017, Ying_IROS_2018, Ying_ICRA_2016, Ying_ICRA_2019, Ying_Automatica_2024, Ying_IEEE_TCST_2020}. Namely, using \mbox{quaternion-based} attitude representation and D'Alembert's laws, the \mbox{open-loop} \mbox{state-space} dynamics of the system can be written as
\begin{align}
\begin{split}
\bs{\dot{\qbar}} &= \frac{1}{2} \bs{\qbar}\otimes
\begin{bmatrix}
0 \\
\bs{\omega} 
\end{bmatrix},\\
\bs{\dot{\omega}} &=  \bs{J}^{-1}\left(\bs{\tau}-\bs{\omega}\times \bs{J}\bs{\omega}\right),
\label{EQ01}
\end{split}
\end{align}
in which $\bs{\qbar}$ is a unit quaternion that stores the attitude of $\bs{\mathcal{B}}$ relative to $\bs{\mathcal{N}}$; $\bs{\omega}$ is the angular velocity of $\bs{\mathcal{B}}$ relative to $\bs{\mathcal{N}}$, written in $\bs{\mathcal{B}}$; $\bs{J}$ is the inertia matrix of the robot written in $\bs{\mathcal{B}}$; and $\bs{\tau}$ is the \mbox{aerodynamic-torque} input signal computed by the controller. As it is well known~\cite{KuipersQuaternions, markley2014fundamentals, AltmannRotQuat}, 
\begin{align}
\bs{\qbar} = 
\left[
\begin{array}{c}
~\cos \frac{\rm{\Theta}}{2} \\
\vspace{-2ex}
\\
\bs{u} \sin \frac{\rm{\Theta}}{2}
\end{array}
\right]\hspace{-0.3ex},
\label{EQ02}
\end{align}
where, the unit vector $\bs{u}$ is the Euler rotation axis and $\rm{\Theta}$ is the rotation amount about $\bs{u}$ required to represent $\bs{\mathcal{B}}$ in space relative to $\bs{\mathcal{N}}$~\cite{KuipersQuaternions}.  

\subsection{Structure of Basic Continuous Controller}
\label{Subsection02B}
For the purpose of controller synthesis and implementation, we define the AEQ as
\begin{align}
\bs{\qbar}_\text{e} = \bs{\qbar}^{-1} \otimes \bs{\qbar}_\text{d},
\label{EQ03}
\end{align}
where \mbox{$\bs{\qbar}_\ts{e} = \left[m_\ts{e}\,\,\bs{n}^T_\ts{e}\right]^T$} represents the attitude of the \mbox{\textit{desired}} \mbox{body-fixed} frame, $\bs{\mathcal{B}}_\ts{d}$, relative to the \textit{actual} \mbox{body-fixed} frame, $\bs{\mathcal{B}}$. Consistently, $\bs{\qbar}$ is the measured instantaneous attitude of $\bs{\mathcal{B}}$ relative to $\bs{\mathcal{N}}$, and $\bs{\qbar}_\ts{d}$ represents the instantaneous attitude of $\bs{\mathcal{B}}_\ts{d}$ relative to $\bs{\mathcal{N}}$. Note that for
\begin{align}
m_\ts{e} = \cos{\frac{\Theta_\ts{e}}{2}}~~~\rm{and}~~~
\bs{n}_{\ts{e}} = \bs{u}_{\ts{e}}\sin \frac{{\Theta}_{\ts{e}}}{2},
\label{EQ04}
\end{align}
$\bs{u}_{\ts{e}}$ is the unit vector about which the robot must rotate an amount ${\Theta}_{\ts{e}}$ to reach $\bs{\mathcal{B}}_\ts{d}$ from $\bs{\mathcal{B}}$. During flight, the control algorithm computes the desired angular velocity of the robot, written in $\bs{\mathcal{B}}_\ts{d}$, as
\begin{align}
\left[
\begin{array}{c}
0 \\
\vspace{-2ex}
\\
\bs{\hat{\omega}}_\text{d}
\end{array}
\right]
 = 2\bs{\qbar}_\ts{d}^{-1} \otimes \bs{\dot{\qbar}}_\text{d}.
\label{EQ05}
\end{align}
Then, to obtain the desired angular velocity of the robot in $\bs{\mathcal{B}}$, we use the transformations $\bs{S}_\ts{d}$ and $\bs{S}^T$ as defined in~\cite{BenaMPPC2022}; $\bs{S}_\ts{d}$ transforms vectors from $\bs{\mathcal{B}}_\ts{d}$ to $\bs{\mathcal{N}}$ and $\bs{S}^T$ transforms vectors from $\bs{\mathcal{N}}$ to $\bs{\mathcal{B}}$. Consequently, $\bs{\omega}_\text{d} = \bs{S}^T \bs{S}_\ts{d} \bs{\hat{\omega}}_\text{d}$. Last, the \mbox{aerodynamic-torque} input is computed according to the continuous control law 
\begin{align}
\bs{\tau} = \bs{K}_{\bs{n}} \bs{n}_{\ts{e}} + \bs{K}_{\bs{\omega}} \bs{\omega}_{\ts{e}} + \bs{J} \bs{\dot{\omega}}_{\ts{d}} + \bs{\omega}\times\bs{J}\bs{\omega},
\label{EQ06}
\end{align}
where $\bs{K}_{\bs{n}}$ and $\bs{K}_{\bs{\omega}}$ are \mbox{positive-definite} matrices; $\bs{\omega}_{\ts{d}}$ is the \mbox{angular-velocity} reference for $\bs{\mathcal{B}}$; $\bs{\omega}_{\ts{e}}$ is the \mbox{angular-velocity} tracking error given by \mbox{$\bs{\omega}_{\ts{d}} - \bs{\omega}$}; $\bs{J} \bs{\dot{\omega}}_{\ts{d}}$ is a feedforward term used for faster tracking performance; and, $\bs{\omega}\times\bs{J}\bs{\omega}$ cancels the nonlinear term in the second line of~(\ref{EQ01}).

\subsection{Closed-Loop Fixed Points and Stability Analysis}
\label{Subsection02C}
By differentiating (\ref{EQ03}) and substituting the \mbox{\textit{right-hand side}} (RHS) of (\ref{EQ06}) into the last row of (\ref{EQ01}), we obtain the \mbox{closed-loop} attitude dynamics of the system. Namely,
\begin{align}
\begin{split}
\bs{\dot{\qbar}}_{\ts{e}} &= \frac{1}{2}
\left[
\begin{array}{c}
    0 \\
    \bs{\omega}_{\ts{e}}
\end{array}
\right]
\otimes \bs{\qbar}_{\text{e}},\\
    \bs{\dot{\omega}}_{\ts{e}} 
    &= -\bs{J}^{-1} \left( \bs{K}_{\bs{n}} \bs{n}_{\ts{e}} + \bs{K}_{\bs{\omega}} \bs{\omega}_{\ts{e}} \right). \\
\end{split}
\label{EQ07}
\end{align}
This \mbox{state-space} system is autonomous and its state is composed of seven variables. Also, as discussed in~\cite{BenaMPPC2022,Bena2023Yaw}, the vector functions on the RHS of (\ref{EQ07}) are Lipschitz continuous for appropriately chosen references $\bs{\qbar}_\ts{d}$ and $\bs{\omega}_{\ts{d}}$. The two fixed points of \eqref{EQ07} can be found using the method described in~\cite{BenaMPPC2022}; the first corresponds to \mbox{$\bs{\qbar}_{\ts{e}}^{\star} = \left[+1\,\, 0\,\, 0\,\, 0 \right]^T$} and \mbox{$\bs{\omega}_{\ts{e}}^{\star} =  \left[ 0\,\, 0\,\, 0 \right]^T$}, and the second corresponds to \mbox{$\bs{\qbar}_{\ts{e}}^{\dagger} = \left[-1\,\, 0\,\, 0\,\, 0 \right]^T$} and \mbox{$\bs{\omega}_{\ts{e}}^{\star} =  \left[ 0\,\, 0\,\, 0 \right]^T$}. These two equilibria represent the same attitude and \mbox{angular-velocity} tracking errors; therefore, convergence to either of them represents convergence to the same physical orientation. The stability properties of the first fixed point, \mbox{$\{\bs{\qbar}^{\star}_\ts{e}, \bs{\omega}^{\star}_\ts{e}\}$}, can be determined using the method presented in~\cite{BenaMPPC2022,Bena2023Yaw}, where we invoke \textit{Lyapunov's direct method} and \textit{LaSalles's invariance principle}, as stated by \mbox{Theorem\,4.1} and \mbox{Corollary\,4.1} in~\cite{khalil2002nonlinear}.
\vspace{1ex}

\hspace{-2.2ex}\textbf{Proposition\,1.} \textit{Let the attitude and \mbox{angular-velocity} references}, $\bs{\qbar}_\ts{d}$ \textit{and} $\bs{\omega}_\ts{d}$, \textit{be smooth functions of time, and let} $\bs{K}_{\bs{n}}$ \textit{and} $\bs{K}_{\bs{\omega}}$ \textit{be constant \mbox{positive-definite matrices}}. \textit{Then, the fixed point} $\{\bs{\qbar}^{\star}_\ts{e}, \bs{\omega}^{\star}_\ts{e}\}$ \textit{of the closed-loop attitude dynamics specified by (\ref{EQ07}), with} \mbox{$\bs{\qbar}_{\ts{e}}^{\star} = \left[+1\,\, 0\,\, 0\,\, 0 \right]^T$} \textit{and} \mbox{$\bs{\omega}_{\ts{e}}^{\star} =  \left[ 0\,\, 0\,\, 0 \right]^T\hspace{-0.4ex}$}, \textit{is asymptotically stable.}

\vspace{1ex}
\hspace{-2.2ex}\textit{Proof.} See Section\,3.6 in~\cite{BenaMPPC2022}.
\vspace{1ex}

As thoroughly discussed in~\cite{BenaMPPC2022}, the proof of \mbox{Proposition\,1} relies on two logical steps. First, we show that $\{\bs{\qbar}^{\star}_\ts{e}, \bs{\omega}^{\star}_\ts{e}\}$ is locally asymptotically stable; then, we show that $\{\bs{\qbar}^{\dagger}_\ts{e}, \bs{\omega}^{\star}_\ts{e}\}$ is unstable and, therefore, any deviation from it always makes the state of the \mbox{closed-loop} attitude dynamics converge to the stable equilibrium point, $\{\bs{\qbar}^{\star}_\ts{e}, \bs{\omega}^{\star}_\ts{e}\}$. Also, in~\cite{BenaMPPC2022}, we prove the instability of \mbox{$\{\bs{\qbar}^{\dagger}_\ts{e}, \bs{\omega}^{\star}_\ts{e}\}$}, using \textit{Lyapunov's indirect method} as stated in~\mbox{Theorem\,4.7}~of~\cite{khalil2002nonlinear}. Here, we formally state this result through a proposition.
\vspace{1ex}

\hspace{-2.2ex}\textbf{Proposition\,2.} \textit{Let the same conditions of Proposition\,1 apply}. \textit{Then, the fixed point} $\{\bs{\qbar}^{\dagger}_\ts{e}, \bs{\omega}^{\star}_\ts{e}\}$, \textit{with} \mbox{$\bs{\qbar}_{\ts{e}}^{\dagger} = \left[ -1\,\, 0\,\, 0\,\, 0 \right]^T$} \textit{and} \mbox{$\bs{\omega}_{\ts{e}}^{\star} =  \left[ 0\,\, 0\,\, 0 \right]^T\hspace{-0.4ex}$}, \textit{is unstable.}
\vspace{1ex}

\hspace{-2.2ex}\textit{Proof.}
See Appendix\,C in~\cite{BenaMPPC2022}.
\vspace{1ex}

\section{The Performance Problem and Benchmark Controller}
\label{Section03}
As mentioned in~Section\,\ref{Section02}, both CL equilibrium AEQs of the system specified by (\ref{EQ07}), \mbox{$\{\bs{\qbar}^{\star}_\ts{e}, \bs{\omega}^{\star}_\ts{e}\}$} and \mbox{$\{\bs{\qbar}^{\dagger}_\ts{e}, \bs{\omega}^{\star}_\ts{e}\}$}, correspond exactly to the same physical state of the controlled UAV; namely, the condition when the instantaneous position, attitude, and angular velocity of $\bs{\mathcal{B}}$ coincide exactly with those of $\bs{\mathcal{B}}_{\ts{d}}$. However, the first fixed point is stable whereas the second is unstable. This apparent paradox results from the \mbox{quaternion-ambiguity} phenomenon. To see this issue, consider (\ref{EQ04}) and recall that $\bs{u}_{\ts{e}}$ is the unit vector about which $\bs{\mathcal{B}}$ must be rotated an angle $\Theta_{\ts{e}}$ to exactly reach the attitude of $\bs{\mathcal{B}}_{\ts{d}}$. Therefore, \mbox{Propositions\,1~and~2} tell us that when \mbox{$\Theta_{\ts{e}} = 0$} and \mbox{$\bs{\omega}_{\ts{e}}=\bs{0}$}, small deviations from this condition are easily corrected by the system's controller; whereas, when \mbox{$\Theta_{\ts{e}} = 2\pi\,\ts{rad}$} and \mbox{$\bs{\omega}_{\ts{e}}=\bs{0}$}, any deviation from this condition, small or large, induces the system's controller to generate a rotation on the order of $2\pi\,\ts{rad}$. Furthermore, it is straightforward to see that when \mbox{$\pi < \Theta_{\ts{e}} < 2\pi\,\ts{rad}$}, the torque $\bs{K}_{\bs{n}} \bs{n}_{\ts{e}}$ is applied in the direction of the longer path. This dynamic behavior does not pose any problems regarding stability; however, from a performance perspective, it raises numerous energy and \mbox{state-error} optimization research questions. 

In the past, to avoid undesired \mbox{$2\pi$-rad} rotations during the execution of flight maneuvers~\cite{bhat2000topological}, we modified the first term of the control law specified by (\ref{EQ06}) by multiplying it by the sign of \mbox{$m_{\ts{e}}$}~\cite{BenaMPPC2022,Bena2023Yaw,yang2019bee,calderon2019control,Bena_RAL_Perception}. Namely,
\begin{align}
\bs{\tau}_{\ts{b}} = \ts{sgn}\{m_\ts{e}\}\bs{K}_{\bs{n}} \bs{n}_{\ts{e}} + \bs{K}_{\bs{\omega}} \bs{\omega}_{\ts{e}} + \bs{J}\bs{\dot{\omega}}_\ts{d} + \bs{\omega}\times\bs{J}\bs{\omega}. \label{EQ08}
\end{align}
From this point onwards, we refer to (\ref{EQ08}) as the benchmark controller. Clearly, the direction of the torque $\ts{sgn}\{m_\ts{e}\}\bs{K}_{\bs{n}} \bs{n}_{\ts{e}}$ is the same as that of $ \bs{u}_{\ts{e}}$, for \mbox{$ 0 \leq \Theta_{\ts{e}} < \pi$}, and is that of $-\bs{u}_{\ts{e}}$, for \mbox{$ \pi \leq \Theta_{\ts{e}} < 2\pi$}, which ensures that the torque $\ts{sgn}\{m_\ts{e}\}\bs{K}_{\bs{n}}\bs{n}_{\ts{e}}$ is applied in the direction of the shorter rotational path between $\bs{\mathcal{B}}$ and $\bs{\mathcal{B}}_{\ts{d}}$. An alternative interpretation of the switching method specified by (\ref{EQ08}) is that when $\ts{sgn}\{m_\ts{e}\}$ becomes negative, \mbox{$\{\bs{\qbar}^{\dagger}_\ts{e}, \bs{\omega}^{\star}_\ts{e}\}$} becomes asymptotically stable and \mbox{$\{\bs{\qbar}^{\star}_\ts{e}, \bs{\omega}^{\star}_\ts{e}\}$} unstable. This stability property of the system is reversed when $\ts{sgn}\{m_\ts{e}\}$ switches back to positive. There is ample evidence that this method is highly effective and robust in \mbox{real-time} flight applications~\cite{Bena2023Yaw,BenaMPPC2022,yang2019bee, calderon2019control,Bena_RAL_Perception}. However, it does not guarantee the best performance regarding the control effort, which is directly related to energy consumption. We hypothesize that replacing $\ts{sgn}\{m_\ts{e}\}$ with a function that predicts the most \mbox{cost-efficient} direction of the first term in the torque control input might be a better option. This is the motivation for introducing the \mbox{MPS-based} controller discussed next. 

\section{Model Predictive Selection}
\label{Section04}
To our best knowledge, all the published methods that select the stable CL equilibrium AEQ during flight use the current value of the system state only~\cite{Mayhew_Robust,QuatAutomatica}. Here, we introduce the MPS algorithm, an alternative approach that predicts and selects the most \mbox{cost-efficient} stable CL equilibrium AEQ over a \mbox{finite-time} horizon, according to a PFM. In this case, we use the PFM
\begin{align}
    \Gamma(t) = \int^{t+t_\ts{h}}_t \left[ \bs{\tau}_{\hspace{-0.2ex}\sigma}^T(\zeta)\bs{R}\bs{\tau}_{\hspace{-0.2ex}\sigma}(\zeta) + \bs{n}_\ts{e}^T(\zeta)\bs{Q}\bs{n}_\ts{e}(\zeta)\, \right] d\zeta,
     \label{EQ09}
\end{align}
where $t_\ts{h}$ is the time horizon; $\bs{\tau}_{\hspace{-0.2ex}\sigma}$ is the torque exciting the system, which is generated through a control law that explicitly depends on the parameter $\sigma$; and, $\bs{R}$ and $\bs{Q}$ are Hermitian \mbox{positive-definite} weight matrices that function as tuning parameters. We specify the control law as  
\begin{align}
\bs{\tau}_{\hspace{-0.2ex}\sigma} = \sigma\bs{K}_{\bs{n}} \bs{n}_\ts{e}  + \bs{K}_{\bs{\omega}} \bs{\omega}_{\ts{e}} + \bs{J}\bs{\dot{\omega}}_{\ts{d}} + \bs{\omega}\times\bs{J}\bs{\omega},
\label{EQ10}
\end{align}
where \mbox{$\sigma \in \{-1,+1\}$} is dynamically selected according to 
\begin{align}
\sigma^{+} =
\left\{
\begin{array}{c}
\hspace{-1.8ex}\,\,~~\sigma,~~~\,\,\ts{if}~~~\, -\delta < \Delta\Gamma < \delta \\
\hspace{-3.0ex}~-1,~~~~\ts{if}~~~~~~~~~~~~ \Delta\Gamma \geq \delta   \\
\hspace{-2.5ex}~~\,+1,~~~~\ts{if}~~~~~~~~~~~~ \Delta\Gamma \leq -\delta 
\end{array}
\right.,
\label{EQ11}
\end{align}
where $\sigma$ is the current value, initialized as \mbox{$\sigma = +1$}; \mbox{$\delta > 0$} is a hysteresis margin included to avoid Zeno behavior~\cite{liberzon2003switching}; and, \mbox{$\Delta \Gamma = \Gamma^\star - \Gamma^{\dagger}$}, in which $\Gamma^\star$ and $\Gamma^{\dagger}$ are the PFM values, computed using (\ref{EQ09}), associated with choosing either $\{\bs{\qbar}^\star_\ts{e},\bs{\omega}^\star_\ts{e}\}$ or $\{\bs{\qbar}^{\dagger}_\ts{e},\bs{\omega}^{\star}_\ts{e}\}$ as the stable CL equilibrium AEQ, respectively. 
\begin{figure}[t!]
\vspace{1.2ex}
\begin{center}
\includegraphics[width=0.48\textwidth]{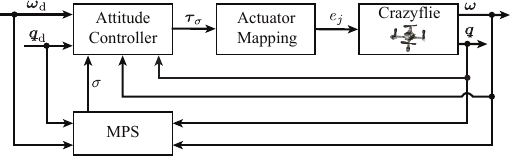}
\end{center}
\vspace{-2ex}
\caption{\textbf{Block Diagram of the \mbox{MPS-based} Control Scheme.} The MPS algorithm receives as inputs the \mbox{attitude-reference} quaternion, measured quaternion, \mbox{angular-velocity} reference, and measured angular velocity; then, it selects the value of $\sigma$ and sends it to the attitude controller, which computes the \mbox{aerodynamic-torque} control input, \mbox{$\bs{\tau}_{\hspace{-0.2ex}\sigma}$}. Last, the actuator mapping receives as input \mbox{$\bs{\tau}_{\hspace{-0.2ex}\sigma}$} and maps it into the \textit{pulse width modulation} (PWM) signals, $e_j$, for $j\in\{1,2,3,4\}$, that excite the DC motors of the flyer as described in~\cite{Ying_ICRA_2016}. \label{Fig02}}
\vspace{-2ex}    
\end{figure}

Since the possible trajectories of $\bs{\tau}_{\hspace{-0.2ex}\sigma}(t)$ and $\bs{n}_{\ts{e}}(t)$, for $t \in \left[t : t + t_{\ts{h}} \right]$, are unknown at time $t$, in order to compute $\Gamma^\star$ and $\Gamma^{\dagger}$, we estimate them using the model specified by (\ref{EQ01})--(\ref{EQ07}). Specifically, using (\ref{EQ01})--(\ref{EQ07}) and (\ref{EQ10}), we simulate the trajectory of the UAV during the time range specified by a \mbox{finite-time} horizon, $t_{\ts{h}}$, for both \mbox{$\sigma=+1$} and \mbox{$\sigma=-1$}, and then select the stable CL equilibrium AEQ associated with the lesser cost of performing the corresponding maneuver, according to the PFM specified by (\ref{EQ09}). The prediction algorithm is based on the methods in Chapter\,12 of~\cite{borrelli_bemporad_morari_2017} and we briefly describe them here. First, since the algorithm is implemented on a \textit{digital signal processor} (DSP), we define \mbox{$\bs{x}(i) = \left[ \bs{\qbar}_\ts{e}^T(i)~\bs{\omega}_{\ts{e}}^T(i) \right]^T$}, where $\bs{\qbar}_\ts{e}(i)$ and $\bs{\omega}_{\ts{e}}(i)$ are the sampled versions of $\bs{\qbar}_\ts{e}(t)$ and $\bs{\omega}_{\ts{e}}(t)$, respectively. Then, we define $\bs{x}_{i+k|i}$ as the state at discrete time \mbox{$i + k$} predicted at discrete time $i$, with the starting point \mbox{$\bs{x}_{i|i} = \bs{x}(i)$}. Next, we solve \mbox{$\bs{x}_{i+k+1|i} = \bs{f}(\bs{x}_{i+k|i})$}, with \mbox{$k \in \left[0 : i_{\ts{h}} -1 \right]$}, for the sequences $\bs{\qbar}_\ts{e}(i)$ and $\bs{\omega}_\ts{e}(i)$, with $i \in \left[i:i+ i_{\ts{h}}\right]$, where $i_{\ts{h}}$ is the \mbox{discrete-time} horizon corresponding to the continuous $t_{\ts{h}}$ and $\bs{f}(\bs{x})$ denotes the sampled version of the RHS vector function in~\eqref{EQ07}. At each time instant, we predict the evolution of the system state for the two possible inputs, $\bs{\tau}_{+1}$ and $\bs{\tau}_{-1}$, associated with choosing either \mbox{$\{\bs{\qbar}^\star_\ts{e},\bs{\omega}^\star_\ts{e}\}$} or \mbox{$\{\bs{\qbar}^{\dagger}_\ts{e},\bs{\omega}^{\star}_\ts{e}\}$} as the stable CL equilibrium AEQ. Last, we use these predictions to execute the selection rule specified by (\ref{EQ11}). 

The \mbox{MPS-based} control scheme is shown in~Fig.\,\ref{Fig02}. We do not provide a rigorous and detailed stability proof of this scheme here; however, we state some basic facts regarding this matter. The inclusion of the hysteresis margin, $\delta$, in the definition of the switching law specified by (\ref{EQ11}) prevents the system from becoming unstable due to chattering. Once chattering is avoided, as long as the responses of the two \mbox{closed-loop} subsystems corresponding to \mbox{$\sigma = -1$} and \mbox{$\sigma=+1$} approach zero as time increases, the switching \mbox{closed-loop} system corresponding to the control law specified by (\ref{EQ10}), as a whole, remains stable. This conclusion follows from noticing that each switching event can be thought of as the reinitialization of one of the two subsystems with a finite initial condition. The experimental results presented in Section\,\ref{Section05} provide strong evidence supporting this statement.
\begin{figure}[t!]
\vspace{1.2ex}
\begin{center}
\includegraphics[width=0.48\textwidth]{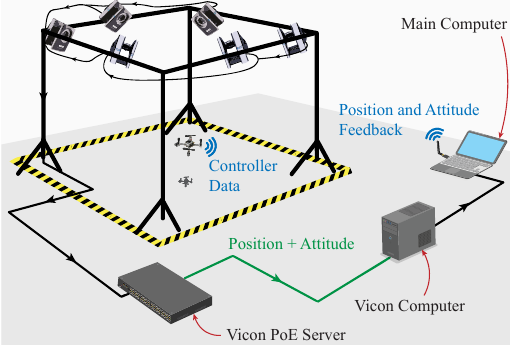}
\end{center}
\vspace{-2ex}
\caption{\textbf{Experimental setup used in the flight tests.} Illustration of the flying arena and \mbox{signals-and-systems} diagram. The setup is equipped with a \mbox{six-V5-camera} Vicon \mbox{motion-capture} system, which is used to measure the position and attitude of the UAV during flight at a rate of $500\,\ts{Hz}$. These data are then sent, via the Vicon~PoE server, to the Vicon computer, which records and transmits the data to the main computer using the \textit{virtual-reality peripheral network} (VRPN) interface. Last, the main computer transmits the position and attitude \mbox{real-time} data to the UAV, using the \mbox{Python-Crazyflie-Client} library, via radio at approximately $50$\,Hz. The UAV transmits the signals generated by the online
controller to the main computer using radio communication. 
\label{Fig03}}
\vspace{-2ex}
\end{figure}

\section{Real-Time Flight Experiments}
\label{Section05}
\subsection{Experimental Setup}
\label{Subsection05A}
The flying arena used in the flight tests discussed in~\mbox{Section\,\ref{Subsection05B}} is depicted in~Fig.\,\ref{Fig03}. This setup is instrumented with a \mbox{six-V5-camera} Vicon \mbox{motion-capture} system running at $500$\,Hz. These cameras measure the position and attitude of the UAV during flight. Then, these data are fused with the \mbox{angular-velocity} signal, which is measured with an onboard BMIO88 \textit{inertial measurement unit}~(IMU) sensor, using the extended K\'alm\'an filter in~\cite{MuellerHamerUWB2015,MuellerCovariance2016}. Specifically, the position and attitude of the UAV are sent to a processing server (\mbox{Vicon~PoE}) and, then, to the Vicon computer that runs Tracker\,3.9 (software). The Vicon computer communicates with the main computer of the setup through the \textit{virtual-reality peripheral network} (VRPN) protocol. Last, the main computer transmits the Vicon measured data to the UAV via radio, using a Bitcraze antenna (Crazyradio~PA) and the \mbox{Python-Crazyflie-Client} library at approximately $50\,\ts{Hz}$. The UAV transmits the signals generated by the online controller to the main computer using radio communication. Both the proposed \mbox{MPS-based} and benchmark controllers are run onboard at a rate of $500\,\ts{Hz}$. During flight, the position of the UAV is controlled using the scheme described in~\mbox{\cite{BenaMPPC2022, Bena2023Yaw}}.
\begin{figure}[t!]
\vspace{1.2ex}
\begin{center}
\includegraphics[width=0.48\textwidth]{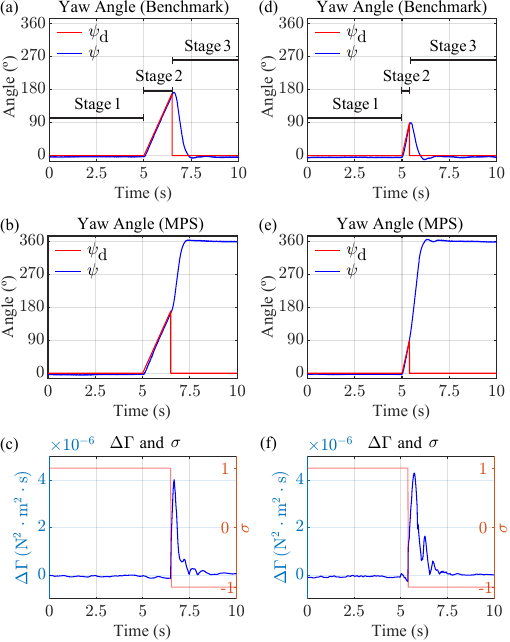}
\end{center}
\vspace{-2ex}
\caption{\textbf{Experimental signals corresponding to two different attitude references and two different controllers.} \textbf{(a)}~\mbox{Real-time} yaw reference and measured yaw signal for the parameter pair \mbox{$\{2\cdot\bs{b}_3\,\ts{rad}\cdot\ts{s}^{-1},170^\circ\}$} and benchmark controller. \textbf{(b)}~\mbox{Real-time} yaw reference and measured yaw signal for the parameter pair \mbox{$\{2\cdot\bs{b}_3\,\ts{rad}\cdot\ts{s}^{-1},170^\circ\}$} and \mbox{MPS-based} controller. \textbf{(c)}~\mbox{Real-time} $\Delta\Gamma$ and $\sigma$ signals corresponding to the case in (b), i.e., \mbox{$\{2\cdot\bs{b}_3\,\ts{rad}\cdot\ts{s}^{-1},170^\circ\}$}. \textbf{(d)}~\mbox{Real-time} yaw reference and measured yaw signal for the parameter pair \mbox{$\{4\cdot\bs{b}_3\,\ts{rad}\cdot\ts{s}^{-1},90^\circ\}$} and benchmark controller. \textbf{(e)}~\mbox{Real-time} yaw reference and measured yaw signal for the parameter pair \mbox{$\{4\cdot\bs{b}_3\,\ts{rad}\cdot\ts{s}^{-1},90^\circ\}$} and \mbox{MPS-based} controller. \textbf{(f)}~\mbox{Real-time} $\Delta\Gamma$ and $\sigma$ signals corresponding to the case in (e), i.e., \mbox{$\{4\cdot\bs{b}_3\,\ts{rad}\cdot\ts{s}^{-1},90^\circ\}$}. \label{Fig04}}
\vspace{-2ex}    
\end{figure}
\begin{figure*}[t!]
\vspace{1.2ex}
\begin{center}
\includegraphics[width=0.99\textwidth]{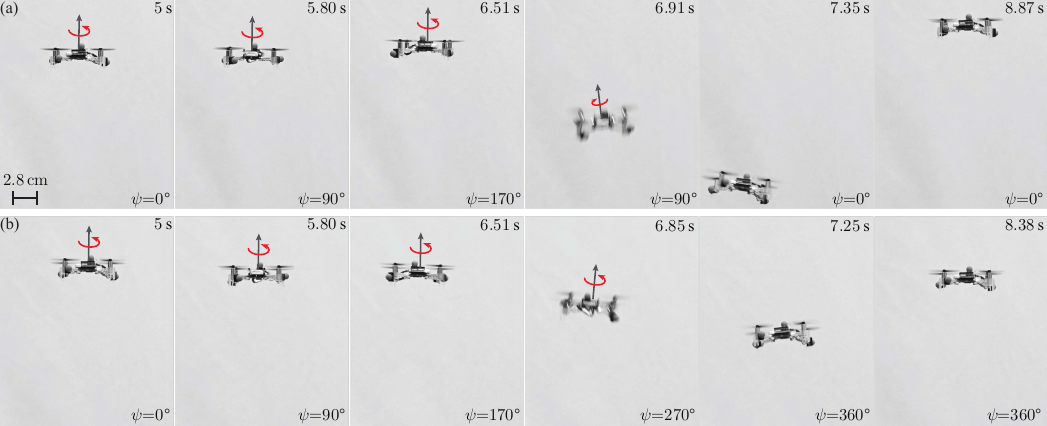}
\end{center}
\vspace{-2ex}
\caption{\textbf{Two flight tests performed using the \mbox{MPS-based} controller and benchmark scheme.} \textbf{(a)}~Photographic sequence from video footage of a flight test performed using the benchmark controller and the attitude reference defined by the parameter pair \mbox{$\{2\cdot\bs{b}_3\,\ts{rad}\cdot\ts{s}^{-1},170^\circ\}$}. The third frame corresponds to the instant when the yaw reference is set to $0^\circ$; in this case, the benchmark controller selects \mbox{$\sigma=+1$}, which produces a reverse rotation and performance degradation. \textbf{(b)}~Photographic sequence from video footage of a flight test performed using the proposed \mbox{MPS-based} controller and the attitude reference defined by the parameter pair \mbox{$\{2\cdot\bs{b}_3\,\ts{rad}\cdot\ts{s}^{-1},170^\circ\}$}. The third frame corresponds to the instant when the yaw reference is set to $0^\circ$; in this case, the \mbox{MPS-based} method selects \mbox{$\sigma=-1$}, which maintains the current direction of rotation and, thus, leads to high flight performance. These two experiments can be viewed in the accompanying supplementary movie.
\label{Fig05}}
\vspace{-1ex}
\end{figure*}
\begin{figure}[t!]
\vspace{1.2ex}
\begin{center}
\includegraphics{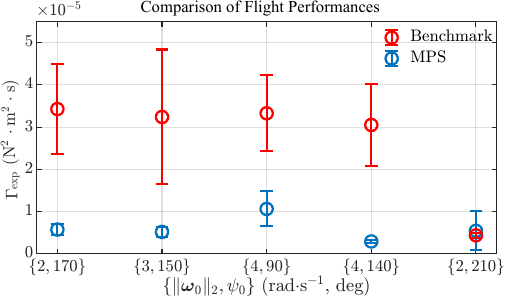}
\end{center}
\vspace{-2ex}
\caption{\textbf{Comparison of flight performances obtained using the benchmark and \mbox{MPS-based} controllers.} Each data point represents the mean and \textit{empirical standard deviation} (ESD) of ten $\Gamma_\ts{exp}$ values corresponding to ten \mbox{back-to-back} flight experiments performed with the same attitude reference and controller. \label{Fig06}}
\vspace{-2ex}
\end{figure}

\subsection{Flight Experiments}
\label{Subsection05B}
To test and demonstrate the capabilities of the proposed \mbox{MPS-based} controller, we performed several different \mbox{high-speed} \mbox{yaw-tracking} maneuvers. The results corresponding to two different \mbox{real-time} attitude references are shown in~Fig.\,\ref{Fig04}. Each reference is composed of three stages. In \mbox{Stage\,1}, the quadrotor hovers stationarily and the rotational commands are set to zero; in \mbox{Stage\,2}, an \mbox{angular-velocity} reference in the $\bs{b}_3$ direction commands the UAV to rotate at the constant value $\bs{\omega}_0$ about its yaw axis; last, in \mbox{Stage\,3}, once the \mbox{yaw-angle} signal $\psi(t)$, with the form of a ramp, reaches a desired \textit{initial} angle \mbox{$\psi(t_0) = \psi_0$}, the constant \mbox{yaw-angle} reference \mbox{$\psi_\ts{d}(t) = 0^\circ$}, for \mbox{$t \geq t_0$}, is inputted to the system, thus compelling the UAV to rotate toward it. Executing pure \mbox{yaw-tracking} maneuvers allows us to straightforwardly evaluate the functionality and performance of different control methods. For the \mbox{real-time} implementation of the controllers, we selected \mbox{$\bs{K}_{\bs{n}} = 900\cdot \bs{J}\,~\ts{N} \cdot \ts{m}$} and \mbox{$\bs{K}_{\bs{\omega}} = 90\cdot \bs{J}\,~\ts{N} \cdot  \ts{m} \cdot \ts{s} \cdot \ts{rad}^{-1}$} as the controller gains in (\ref{EQ08}) and (\ref{EQ10}). Additionally, we empirically selected the \mbox{finite-time} horizon for the \mbox{MPS-based} controller to be $0.4\,\ts{s}$, and the tunable weights in \eqref{EQ09} to be \mbox{$\bs{R} = \bs{I}$} and \mbox{$\bs{Q} = 10^{-6}\cdot\bs{I}\,~\ts{N}^2\cdot\ts{m}^2$}, where $\bs{I}$ is the identity matrix with compatible dimensions. The reason for choosing significantly lower values for the diagonal entries of $\bs{Q}$ is the difference in order of magnitude between \mbox{$\bs{\tau}_{\hspace{-0.2ex}\sigma}^T\bs{\tau}_{\hspace{-0.2ex}\sigma}$} and \mbox{$\bs{n}_\ts{e}^T\bs{n}_\ts{e}$}. Last, we set the hysteresis margin to be \mbox{$\delta = 5\cdot10^{-7}$}, which was empirically determined to be large enough to avoid chattering.

The data in \mbox{Figs.\,\ref{Fig04}(a)-(c)} correspond to an attitude reference defined by the pair \mbox{$\left\{ \bs{\omega}_0 = 2 \cdot \bs{b}_3~\ts{rad} \cdot \ts{s}^{-1}, \psi_0 = 170^{\circ} \right\}$}. Fig.\,\ref{Fig04}(a) shows the reference $\psi_{\ts{d}}$ and measurement $\psi$, when the UAV is flown by the benchmark controller. Fig.\,\ref{Fig04}(b) shows the reference $\psi_{\ts{d}}$ and measurement $\psi$, when the UAV is flown by the \mbox{MPS-based} controller. For this value of $\psi_0$, the \mbox{MPS-based} controller forces the UAV to continue rotating in the direction of $\bs{u}_{\ts{e}}$ and not reverse its rotation when the attitude reference drastically changes, whereas the benchmark controller forces the UAV to reverse its direction of rotation when the attitude reference drastically changes. Fig.\,\ref{Fig04}(c) shows the time evolution of signals $\Delta \Gamma$ and $\sigma$ corresponding to the data in~Fig.\,\ref{Fig04}(b), obtained with the \mbox{MPS-based} controller. Since the controllers are implemented on a DSP, to evaluate and compare the flight performances achieved with the \mbox{MPS-based} and benchmark controllers, we use a \mbox{discrete-time} version of the PFM specified by (\ref{EQ09}), over the time interval during which the maneuver is executed. Namely,
\begin{align}
{\small
\Gamma_\ts{exp} = \sum_{i=1}^{N_{\ts{s}}} \left[\bs{\tau}_{\ts{exp}}^T(i)\bs{R}\bs{\tau}_{\ts{exp}}(i) + \bs{n}_{\ts{e},\ts{exp}}^T(i)\bs{Q}\bs{n}_{\ts{e},\ts{exp}}(i)\right]\cdot T_\ts{s},
}
\label{EQ12}
\end{align}
where $N_{\ts{s}}$ is the number of samples collected during the flight maneuver and $T_{\ts{s}}$ is the sampling time. In all cases discussed here, \mbox{$N_{\ts{s}} = 1500$} and \mbox{$T_{\ts{s}} = 0.002\,\ts{s}$}. The subscript `exp' simply means `experimental.' For the cases in \mbox{Figs.\,\ref{Fig04}(a)-(c)}, the experimental PFM for the benchmark controller is \mbox{$2.78 \times 10^{-5}\,\ts{N}^2 \cdot \ts{m}^2 \cdot \ts{s}$}, and for the \mbox{MPS-based} controller is \mbox{$6.25 \times 10^{-6}\,\ts{N}^2 \cdot \ts{m}^2 \cdot \ts{s}$}. Video footage of the two flight tests corresponding to \mbox{Figs.\,\ref{Fig04}(a)~and~(b)} are shown in the accompanying supplementary movie. Also, Fig.\,\ref{Fig05} shows photographic sequences of these two experiments. Here, the superior performance of the \mbox{MPS-based} controller over the benchmark scheme is evident.

The data in \mbox{Figs.\,\ref{Fig04}(d)-(f)} correspond to an attitude reference defined by the pair \mbox{$\left\{ \bs{\omega}_0 = 4 \cdot \bs{b}_3~\ts{rad} \cdot \ts{s}^{-1}, \psi_0 = 90^{\circ} \right\}$}. Fig.\,\ref{Fig04}(d) shows the reference $\psi_{\ts{d}}$ and measurement $\psi$, when the UAV is flown by the benchmark controller. Fig.\,\ref{Fig04}(e) shows the reference $\psi_{\ts{d}}$ and measurement $\psi$, when the UAV is flown by the \mbox{MPS-based} controller. For this value of $\psi_0$, the \mbox{MPS-based} controller forces the UAV to continue rotating in the direction of $\bs{u}_{\ts{e}}$ and not reverse its rotation when the attitude reference drastically changes, whereas the benchmark controller forces the UAV to reverse its rotation when the attitude reference drastically changes. Fig.\,\ref{Fig04}(f) shows the time evolution of signals $\Delta \Gamma$ and $\sigma$ corresponding to the data in~Fig.\,\ref{Fig04}(e), obtained with the \mbox{MPS-based} controller. For the cases in \mbox{Figs.\,\ref{Fig04}(d)-(f)}, the experimental PFM for the benchmark controller is \mbox{$4.82 \times 10^{-5}\,\ts{N}^2 \cdot \ts{m}^2 \cdot \ts{s}$}, and for the \mbox{MPS-based} controller is \mbox{$1.29 \times 10^{-5}\,\ts{N}^2 \cdot \ts{m}^2 \cdot \ts{s}$}.

Last, Fig.\,\ref{Fig06} summarizes ten sets of experimental results obtained during Stage\,3 of flight tests. Each experimental result in a set corresponds to a pair of \mbox{attitude-reference} parameters, $\left\{\bs{\omega}_0, \psi_0 \right\}$, and a type of controller. Each data point in the plot of Fig.\,\ref{Fig06} represents the mean and \textit{empirical standard deviation} (ESD) of ten $\Gamma_\ts{exp}$ values corresponding to ten \mbox{back-to-back} flight experiments performed with the same reference and control system. The data collected using the benchmark controller are shown in red and those collected using the \mbox{MPS-based} controller are shown in blue. These data show that, on average, the \mbox{MPS-based} controller reduces the value of $\Gamma_\ts{exp}$ by $60.30\,\%$ when compared to the benchmark controller. Furthermore, in all the cases in which the \mbox{MPS-based} and benchmark controllers select different CL equilibrium AEQs, the worst performance obtained with the \mbox{MPS-based} controller is better than the best performance achieved with the benchmark controller. This comparison compellingly demonstrates the superior performance of the proposed \mbox{MPS-based} controller. In the last case, corresponding to the pair \mbox{$\left\{ \bs{\omega}_0 = 2 \cdot \bs{b}_3~\ts{rad} \cdot \ts{s}^{-1}, \psi_0 = 210^{\circ} \right\}$}, the performances achieved by both controllers are very similar because they select the same stable CL equilibrium AEQ. It is worth noting that the ESD values for all \mbox{benchmark-controller} data points are significantly larger than those of the \mbox{MPS-based-controller} data points. This observation is explained by the reverse rotations forced by the former method that the latter method avoids; reverse rotations degrade the flight performance and make repeatability almost impossible.

\vspace{0.4ex}
\section{Conclusions}
\label{Section06}
We presented a new attitude control method that dynamically selects the stable CL equilibrium AEQ of a UAV during \mbox{high-speed} \mbox{yaw-rotational} flight. The main objective of this approach is minimizing the cost of flight according to a PFM that explicitly accounts for both the control input and \mbox{attitude-error} state of the UAV. The method is based on MPS, a new algorithm implemented in real time to estimate the cost of flight over a \mbox{finite-time} horizon, using a dynamic model derived from first principles. To test and demonstrate the functionality and performance of the proposed method, we performed one hundred \mbox{yaw-tracking} flight tests using an \mbox{MPS-based} controller and a benchmark scheme. For these tests, the experimental data show that the \mbox{MPS-based} controller reduces the value of the \mbox{cost-of-flight} PFM by $60.30\,\%$, on average, with respect to that obtained with the benchmark controller. These results highlight the suitability and potentials of the \mbox{MPS-based} approach.

\newpage

\balance
\bibliographystyle{IEEEtran}
\bibliography{paper}

\end{document}